\documentclass[10pt,twocolumn,letterpaper]{article}

\usepackage{wacv}              

\usepackage[accsupp]{axessibility} 

\usepackage{graphicx}
\usepackage{amsmath}
\usepackage{amssymb}
\usepackage{booktabs}
\usepackage{verbatim}
\usepackage{tablefootnote}
\usepackage{subcaption} 

\usepackage[pagebackref,breaklinks,colorlinks]{hyperref}

\usepackage[capitalize]{cleveref}
\crefname{section}{Sec.}{Secs.}
\Crefname{section}{Section}{Sections}
\Crefname{table}{Table}{Tables}
\crefname{table}{Tab.}{Tabs.}

\def\wacvPaperID{*****} 
\def\confName{WACV}
\def\confYear{2025}

\begin{document}

\title{MCTR: Multi Camera Tracking Transformer}

\author{Alexandru Niculescu-Mizil\\
{\tt\small alexnic@gmail.com}
\and
Deep Patel, Iain Melvin\\
NEC Laboratories America\\
{\{dpatel, iain\}@nec-labs.com}
}

\maketitle

\begin{abstract}
Multi-camera tracking plays a pivotal role in various real-world applications. While end-to-end methods have gained significant interest in single-camera tracking, multi-camera tracking remains predominantly reliant on heuristic techniques. In response to this gap, this paper introduces Multi-Camera Tracking tRansformer (MCTR), a novel end-to-end approach tailored for multi-object detection and tracking across multiple cameras with overlapping fields of view. MCTR leverages end-to-end detectors like DEtector TRansformer (DETR) to produce detections and detection embeddings independently for each camera view. The framework maintains set of track embeddings that encaplusate global information about the tracked objects, and updates them at every frame by integrating the local information from the view-specific detection embeddings. The track embeddings are probabilistically associated with detections in every camera view and frame to generate consistent object tracks. The soft probabilistic association facilitates the design of differentiable losses that enable end-to-end training of the entire system. To validate our approach, we conduct experiments on MMPTrack and AI City Challenge, two recently introduced large-scale multi-camera multi-object tracking datasets. 
Code is available at \url{https://github.com/necla-ml/mctr}.
\end{abstract}

\section{Introduction}
\label{sec:intro}

Object tracking has long been a central challenge in computer vision attracting substantial attention in the research community due to its applicability in numerous real-world applications. While the majority of the research efforts have concentrated on multi-object tracking in single camera video feeds, there has been a rising demand for multi-object multi-camera tracking due to the increasing prevalence of multi-camera systems deployed in diverse applications such as security, monitoring, or sports analytics. In these applications multi-camera setups offer a multitude of advantages over single-camera counterparts, including increased coverage, reduction of blind spots and heightened tracking robustness, particularly in scenarios involving detection failures or prolonged occlusions.  


The predominant approach to multi-camera tracking thus far has revolved around heuristic techniques, which amalgamate various components such as person re-identification, single-camera tracking, homography estimation, and clustering. These techniques have demonstrated commendable performance; however, they are inherently heuristic, and their performance often hinges on the effectiveness of hand-crafted rules and the quality of heuristics used.  Recently, the application of transformer-based algorithms to tracking has received significant attention \cite{sun2020transtrack, zeng2022motr, meinhardt2022trackformer,cai2022memot,zhu2022looking} due to their ability to encapsulate the entirety of the tracking process within a unified, end-to-end framework. However, this transition toward end-to-end tracking primarily applies to single-camera scenarios. In the domain of multi-camera tracking, particularly when dealing with highly overlapping camera views, there has been little work in this direction.

In this paper, we propose the Multi Camera Tracking tRansformer (MCTR), a novel approach to multi-camera multi-object tracking that adopts an end-to-end architecture to track multiple objects across multiple camera feeds. MCTR builds on end-to-end object detection models such as DETR (DEtections TRansformer) \cite{carion2020end}, adding two new components to facilitate multi-camera tracking: a tracking module and an association module. The flow of MCTR is conceptually simple: the object detectors are independently applied on every camera feed to generate view-specific detections and detection embeddings; the tracking module maintains a set of track embeddings and updates them using the information in the view-specific detection embeddings; and the association module generates an assignment of detections to tracks based on the respective detection and track embeddings.  The model is trained end-to-end using a new loss formulation. This approach offers a more principled solution to multi-camera tracking, reducing the reliance on heuristic components and integrating the entire tracking process into a coherent, data-driven framework. 

Multi-camera tracking introduces additional complexity over single-camera tracking as track consistency needs to be maintained both across time and across camera views. This makes it difficult to directly extend single-camera techniques, that generally depend on the assumption that a track may only be associated with a single detection at every time step, like, for example, those employed by MOTR \cite{zeng2022motr} which use the track embeddings as detection queries that are iteratively updated by DETR. Instead, MCTR maintains a separate set of track embeddings that encapsulate global information about the tracked objects across all views, and are distinct from the DETR object embeddings, which encapsulate local, view-specific information. 

As is the case with all transformer-based models, the track embeddings as well as the detection embeddings do not have a particular order, and, thus, do not have a fixed relation with ground truth labels. This makes it challenging to devise a loss for training the model. DETR solves this issue for object detection by first finding the best assignment of detection embeddings to ground truth annotations using the Hungarian algorithm \cite{kuhn1955hungarian}, and then computing classification and detection losses based on this assignment. MOTR and other approaches extend this idea to single-camera tracking by forcing each track query that has been previously associated with a ground truth object to keep detecting the same object in future frames. It is unclear, however, how to extend this procedure to keep track assignments consistent not only across time, but also across camera views. Furthermore, this approach is highly dependent on having a good association between detections and ground truth in the first frame, as it can not be changed in subsequent frames. In this paper we take a different approach to devise a training loss that avoids these issues. To calculate view-specific classification and detection losses we follow DETR's approach and use the Hungarian algorithm in every frame and camera view independently to locally associate detections with ground truth. 
To maintain global tracking consistency, we use an approach akin to the attention mechanism in scaled dot-product attention to generate probabilistic assignments of local detections to global track embeddings.
For pairs of detections in different cameras or different frames we calculate the probability, under the model, that the detections belong to the same track by integrating over the track assignment. We then use a negative log-likelihood loss to train the model to assign high probabilities to detection pairs from the same ground truth track and low probabilities to pairs from different tracks.

In summary, our contributions are: 1) An end-to-end framework for tracking multiple objects in multiple cameras with overlapping fields of view,
2) A method for probabilistic association of detections to tracks that is fully differentiable w.r.t input RGB frames, and 3) The formulation of a specialized loss function designed to guide the model in preserving consistent object identities, both temporally and across diverse camera perspectives.

The primary objective of this study is to test the feasibility of employing transformer-based end-to-end architectures within the realm of multi-camera multi-object tracking problems. Our aim is to demonstrate the potential viability of such techniques and their applicability to these complex scenarios. We anticipate that our findings will inspire further exploration and research in this direction, potentially unlocking the full capabilities of these models in the future.

\begin{figure}
    \centering    \includegraphics[width=\linewidth,height=0.6\linewidth]{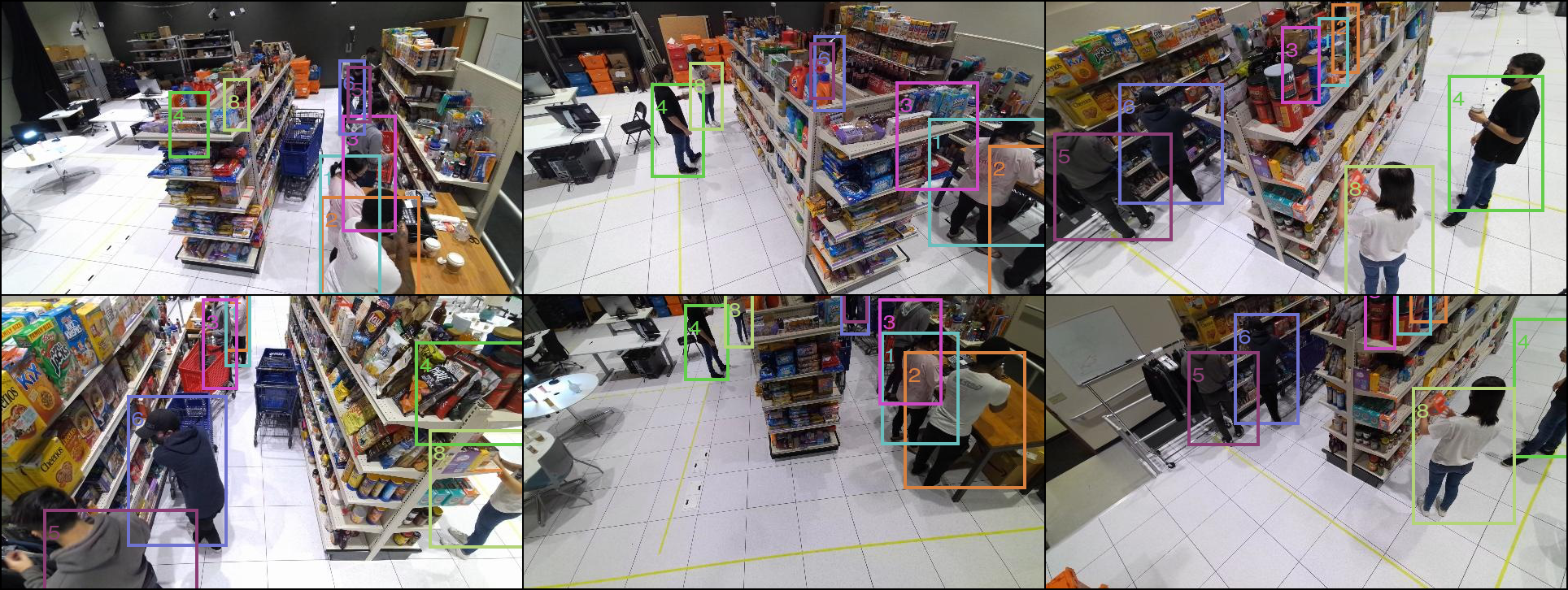}
    \caption{Example of multi-camera frame from the MMPTrack dataset with 6 camera angles.}
    \label{fig:enter-label}
\end{figure}
\section{Related Works}
\label{related_works}

\subsection{Single Camera Tracking}

 Multi-object tracking within a single-camera setup has been extensively studied. Tracking by detection approaches \cite{wojke2017simple, schulter2017deep, zhang2022bytetrack, cao2023observation, leal2016learning} utilize object detectors \cite{duan2019centernet, girshick2015fast, redmon2018yolov3}, to identify objects in individual frames and perform track association using Kalman Filter \cite{welch1995introduction} or Hungarian Matching \cite{kuhn1955hungarian}. Recently, end to end tracking methods such as \cite{sun2020transtrack, zeng2022motr, cai2022memot, zhao2022tracking, zhang2023motrv2, yu2023motrv3, zhu2022looking, meinhardt2022trackformer} have emerged, extending query based object detection for tracking \cite{carion2020end}. For example, MOTR \cite{zeng2022motr}, MOTRv2 \cite{zhang2023motrv2} and TrackFormer \cite{meinhardt2022trackformer} propagate track queries across frames, iteratively updating them with image features for long-term detection and tracking. TransTrack \cite{sun2020transtrack} and P3Aformer \cite{zhao2022tracking} utilize 
 a location-based cost matrix for bipartite matching \cite{kuhn1955hungarian}. In contrast to tracking by detection methods, end-to-end tracking methods are data-driven and avoid dependence on handcrafted heuristics. Despite their effectiveness, single-camera tracking encounters challenges, especially in cluttered and crowded environments with occlusions. The reliance on a single viewpoint limits the system's robustness in complex scenarios. To alleviate this, we extend the tracking by query propagation paradigm for multi-camera multi-object tracking.

\subsection{Multi-Camera Tracking}

The multi-camera tracking (MCT) domain has witnessed diverse methodologies to tackle challenges in complex surveillance environments. A prevalent strategy involves a distributed approach, where single-camera tracking precedes hierarchical clustering \cite{murtagh2012algorithms}, non-negative matrix factorization (NMF) \cite{wang2012nonnegative}, or other merging and association algorithms for tracklets \cite{black2002multi, le2018online, quach2021dyglip, he2020multi, jiang2012multi, specker2021occlusion, yang2022distributed, chen2016equalized, Kohl_2020_CVPR_Workshops, xu2016multi, ristani2018features}. Alternatively, global or centralized methods bypass single-camera tracking, focusing on detecting individuals in each camera view and subsequently globally associating detections into tracklets \cite{yang2022box, liu2017multi}. Occupancy maps derived from multi-camera views have been explored \cite{fleuret2007multicamera, you2020real}, and trajectory prediction is employed in certain works, with variations in global or distributed approaches for multi-view association \cite{yang2022box, Jeon_2023_CVPR}. Iguernaissi et al. \cite{iguernaissi2019people} provides a comprehensive survey of the MCT papers.

Notable contributions in MCT include TRACTA \cite{he2020multi}, utilizing RNMF for cross-camera tracklet matching, and DMCT \cite{you2020real}, incorporating a perspective-aware GroundPoint Network, occupancy heatmap estimation, and a glimpse network for person detection and tracking. Recently, ReST \cite{cheng2023rest} proposed a two-stage association method using a reconfigurable graph model.  In the vision-based autonomous driving systems domain, efforts focus on multi-view 3D object tracking, building upon end-to-end multi-view 3D object detection methodologies \cite{doll2022spatialdetr, wang2022detr3d}. Works such as MUTR3D \cite{zhang2022mutr3d}, PF-Track \cite{pang2023standing}, ViP3D \cite{gu2023vip3d}, and DQTrack \cite{li2023end} utilizes a 3D track query-based approach for coherent object tracking across multiple cameras and frames. DQTrack \cite{li2023end} introduces a decoupled-query paradigm for camera-based 3D multi-object tracking, and ViP3D \cite{gu2023vip3d} pioneers a fully differentiable trajectory prediction approach. PF-Track \cite{pang2023standing} emphasizes spatio-temporal continuity.

In our approach, we conduct global track association for detections in each individual view, departing from single-camera tracklet-to-tracklet association. Diverging from existing global association methods, our fully end-to-end trainable framework based on transformers avoids heuristics or handcrafted pipelines for data-driven learning. Our track association loss is fully differentiable and trained end-to-end, similar to 3D object tracking. However, we focus on 2D person tracking across overlapping cameras.

\section{Method}
\label{method}

\begin{figure}
    \centering
    \includegraphics[width=1.0\linewidth]{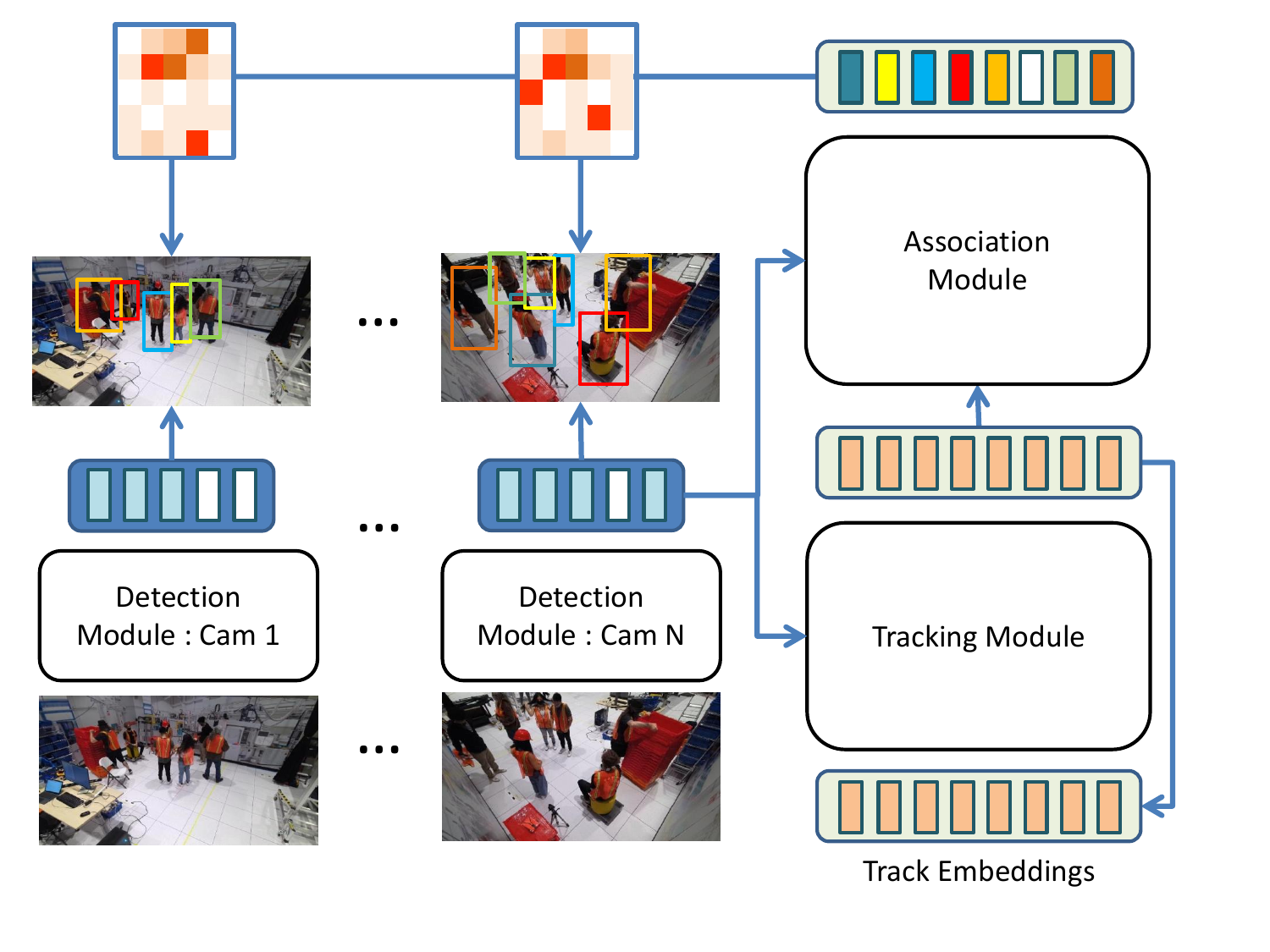}
    \caption{Model Overview.}
    \label{fig:model}
\end{figure}

In this section, we provide a detailed description of the architecture of the proposed model. Figure \ref{fig:model} offers an overview of our model, highlighting its key components and their interactions.

The system consists of a detection module, a tracking module, and an association module. The detection module operates independently for each camera view and generates a set of detection embeddings. These embeddings encapsulate essential information regarding object detections within each individual view and are used to generate a bounding box prediction, a class prediction, and to inform the downstream modules. The tracking module maintains a set of track embeddings. The track embeddings contain global information about each object and are used to maintain consistent and coherent object identities across camera views and time. The track embeddings are updated at each frame using the information in the detection embeddings from all the camera views. The association module is tasked with producing a probabilistic assignment of detections to identities based on the information in the respective track and detection embeddings. 

\subsection{Detection Module}

\begin{figure}
    \centering
    \includegraphics[width=1.0\linewidth]{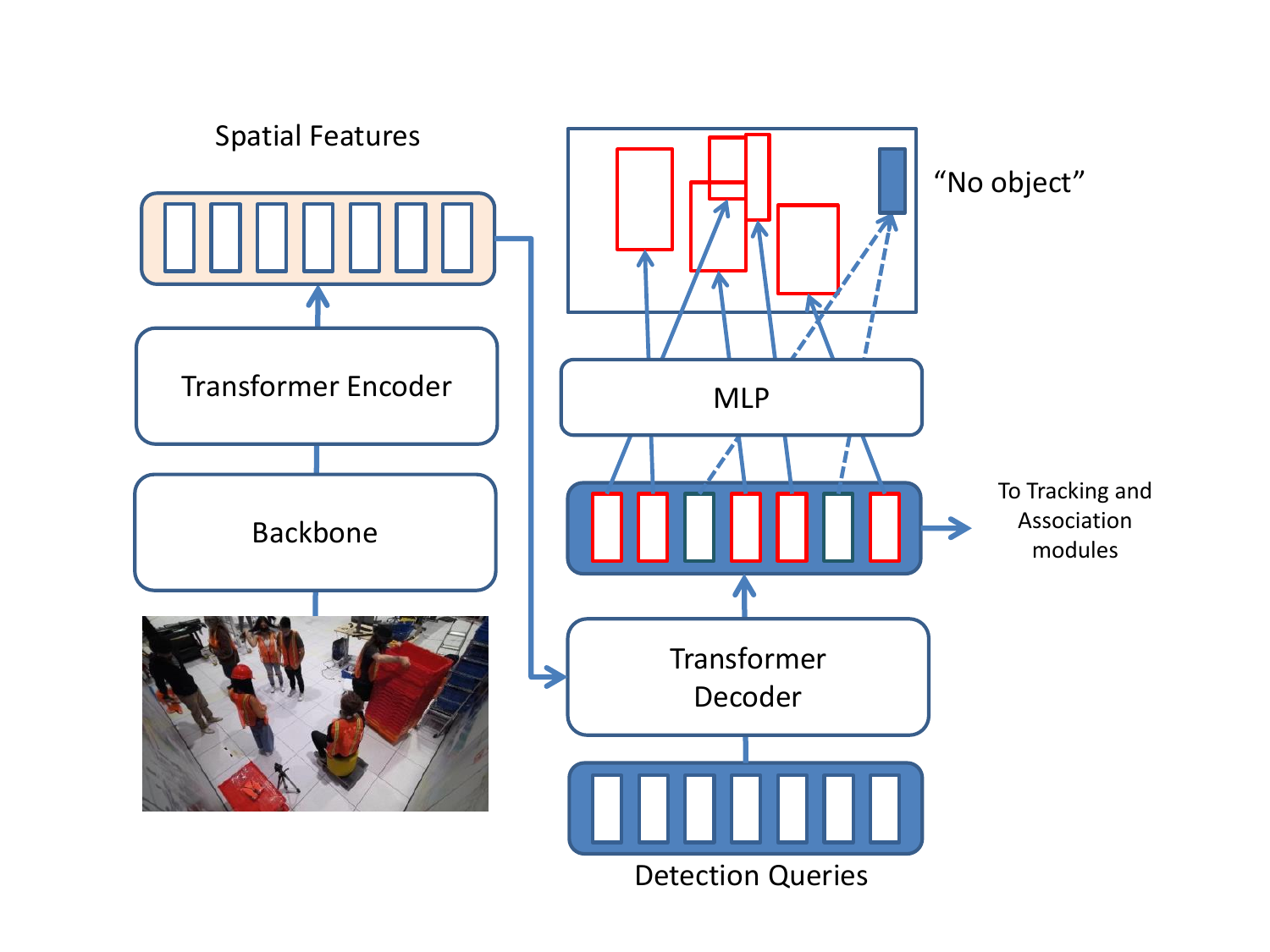}
    \caption{Detection Module: DETR.}
    \label{fig:detection_module}
\end{figure}

In this paper we employ a vanilla DETR (DETection TRansfomer) for the detection module, but any other end-to-end detector architecture can be used.  
A DETR model operates independently on each camera view. Briefly, the architecture of the DETR model includes a backbone to extract image features and a transformer encoder and decoder. The spatial features at the output of the backbone are added to a positional encoding and input to the encoder transformer which performs self attention and outputs a set of spatial features. The DETR transformer decoder takes a set of detection query embeddings as input and outputs a set of detection embeddings. Each of the output embeddings are then passed through an MLP to predict the  object class (or a "no object" class) and its bounding box. An overview of DETR is shown in  figure~\ref{fig:detection_module}.  For more details see \cite{carion2020end}. 

\subsection{Tracking Module}

\begin{figure}
    \centering
        \includegraphics[width=1.0\linewidth]{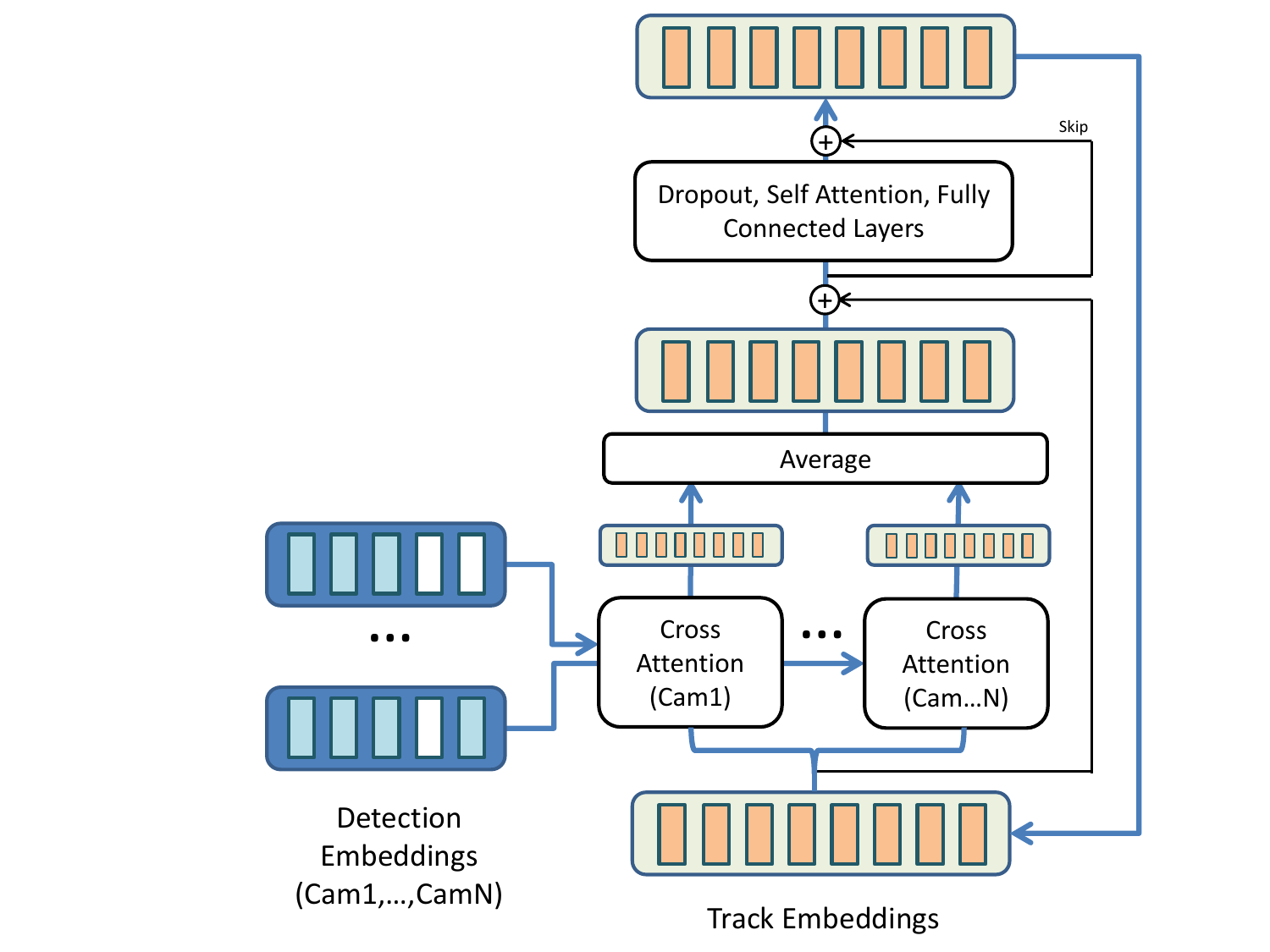}
        \caption{Tracking Module.}
        \label{fig:tracking_module}
\end{figure}

The tracking module is tasked with updating the track embeddings with information from all the camera views in the current frame. The purpose of the track embeddings is to maintain global information about the tracked objects. The architecture of the tracking module is depicted in figure~\ref{fig:tracking_module}. 

The track embeddings are initialised at the first frame to learned "track query embeddings". Like the "query inputs" to DETR,  some of the embeddings will take on the role of representing one identity over time and others will remain unassigned. When a track embedding remains unassigned to a detection after a certain time, it is reset to it's initial embedding.
To update the track embeddings with detections from each camera, the cross-attention section has a set of multi-head cross-attention modules tailored to each camera view, with each module computing cross-attention between the current track embeddings and the detection embeddings from the corresponding view. 
The track embeddings act as "queries", and detection information as "keys" and "values". 
Because the relationship between objects and tracks depends on the camera position, each cross-attention module has its own distinct parameters (i.e. there is no parameter sharing between the cross-attention modules corresponding to the different views). The outputs of all the view specific cross-attention modules are averaged and passed through self-attention and feed-forward layers to obtain updated track embeddings. 

The self-attention module in the tracking module is intended to introduce competition between the track embeddings. If one track query 'claims' a particular identity strongly, other track queries can "observe" this and are encouraged not to take on the same identity through the attention mechanism. This is similar to the way self attention works in the decoder for DETR to discourage 2 query embeddings from claiming the same output detection.

\subsection{Association Module}
\label{sec:association}

\begin{figure}
    \centering
    \includegraphics[width=1.0\linewidth]{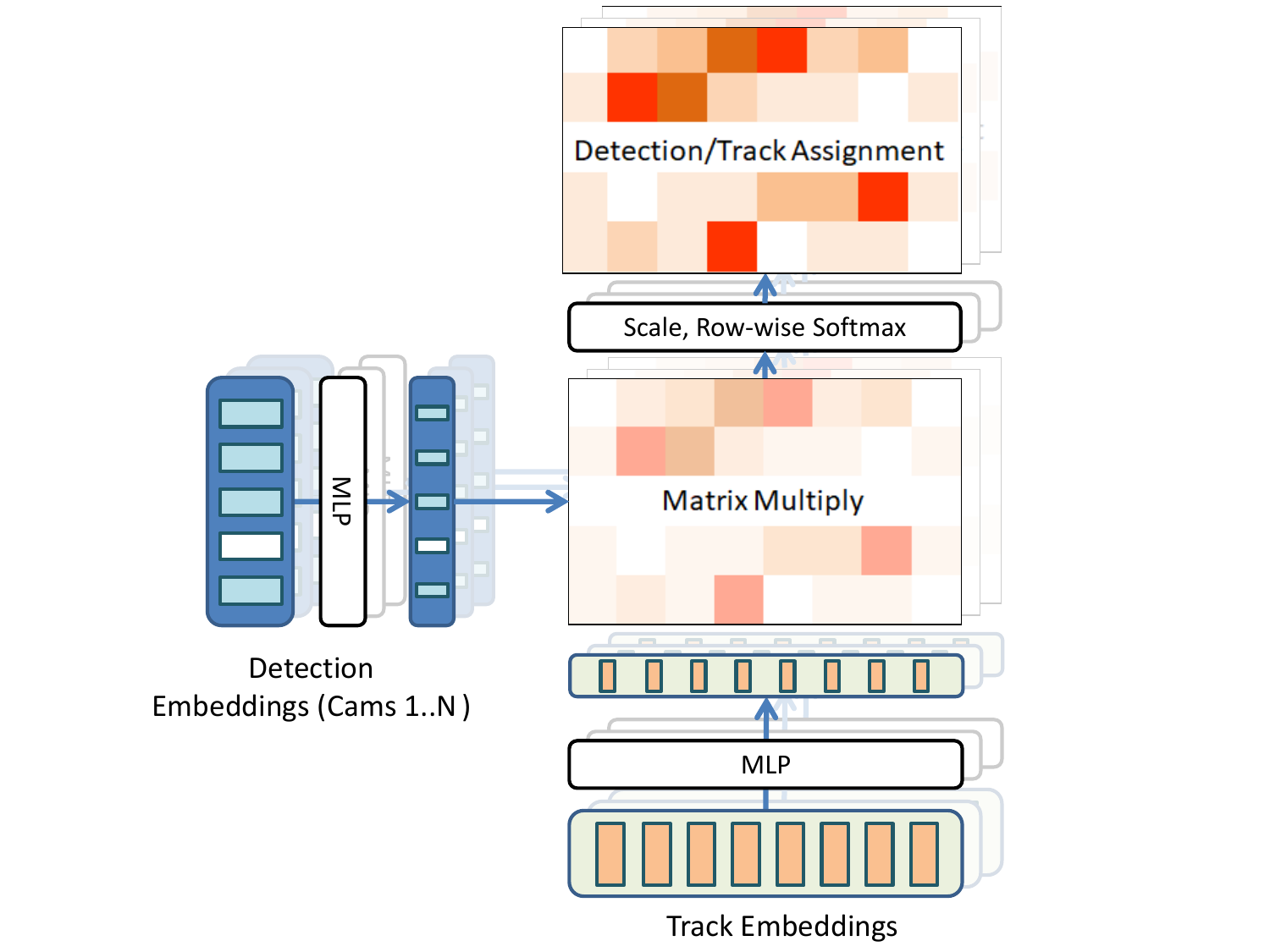}
    \caption{Association Module.}
    \label{fig:association}
\end{figure}

The association module, depicted in figure~\ref{fig:association}, produces a probabilistic assignment of detections to tracks. The assignment is performed independently for each camera view through a mechanism that is the same as the attention mechanism in scaled dot-product attention, wherein the detections act as queries and tracks act as keys. The detection embeddings and track embeddings undergo a linear transformation, followed by the multiplication of the the resulting matrices. Similar to scaled attention, the result is scaled by the square root of the embedding dimension before applying a row-wise softmax operation. The result is a $D \times T$ matrix $A^v$ where each entry $A^{v}_{d,t}$ represents the probability that detection $d$ is associated with track $t$ in view $v$. $D$ is the number of detection queries while $T$ is the number of track queries.   

\subsection{Training Loss}
\label{sec:loss}

To train the model we use several loss functions, that we group into detection losses, track losses and auxiliary track losses. 

{\bf Detection loss. } The detection loss, $\mathcal{L}_{det}$, is the same as the one used used by \cite{carion2020end} to train DETR: a combination of negative log-likelihood classification loss, an IOU loss and a GIOU loss. Given ground truth annotations, Hungarian matching is used to find the bipartite assignment of detections to ground truth that incurs the lowest loss.  This loss is used as a detection loss for each view. Following \cite{carion2020end} we also use losses calculated at different layers of the DETR transformer decoder as auxiliary losses, $\mathcal{L}_{det\_aux}$.

{\bf Track losses. } The track losses enforce consistency of object identity between camera views and camera frames.  Given two detections $d_1$ and $d_2$ from two different views $v_1$ and $v_2$, the probability, according to the model, that $d_1$ and $d_2$ belong to the same track ID can be calculated by integrating over tracks:
\[ 
 P_{st}(d_1, d_2) = \sum_{t} A^{v_1}_{d_1,t} \cdot A^{v_2}_{d_2,t}
\]

where $A^{v_1}$ and $A^{v_2}$ are the probabilistic assignment matrices for views $v_1$ and $v_2$ calculated by the association module. (See section~\ref{sec:association}.) 

To obtain a label, $y_{st}(d_1, d_2)$, for a pair of detections, we leverage the detection to ground truth assignment provided by the Hungarian matching algorithm described above. We have three possible cases: 1) both $d_1$ and $d_2$ are associated with a ground truth annotation, and both ground truth annotations have the same track ID. In this case $y_{st}(d_1, d_2) = 1$; 2) both $d_1$ and $d_2$ are associated with a ground truth annotation, but both ground truth annotations have the different track IDs. In this case $y_{st}(d_1, d_2) = 0$; and 3) either $d_1$ or $d_2$ is not associated with any ground truth annotation. In this case $y_{st}(d_1, d_2)$ is undefined. 

The loss $\mathcal{L}_{accross\_cam}$ is the negative log-likelihood of the labels $y_{st}$, when they are defined:
\begin{align}
  \mathcal{L}_{across\_cams} = & - \frac{1}{N_p}\sum_{d_1,d_2} \left( y_{st}(d_1, d_2) \cdot \log(P_{st}(d_1,d_2)) \right. \nonumber \\
  & + \left. (1 - y_{st}(d_1, d_2)) \cdot \log(1 - P_{st}(d_1,d_2)) \right) \nonumber
\end{align}

where the sum is taken over pairs of detections $d_1$ and $d_2$ from different camera views in the same frame, for which $y_{st}(d_1, d_2)$ is defined. $N_p$ is the number of such pairs. 

Similarly, one can define $\mathcal{L}_{across\_frames}$ by taking pairs of detections from the same camera view and different frames. 

{\bf Auxiliary track losses. } The probabilistic formulation of the assignment of detections to tracks enables the definition of auxiliary losses to encourage learning of more informative track embeddings.  For example one can induce the track embeddings to encode information about the bounding boxes of corresponding object in all camera views.  To achieve this the track embeddings are passed, for each view, through a three layer MLP with ReLU nonlinearity to predict the coordinates $\hat{B}(t,v)$ of the bounding box for the respective track $t$ in view $v$. The MLPs are view specific, since an object position would be different in different camera views. The track IOU loss is: 
\[ 
    \mathcal{L}_{track\_IOU} = \frac{1}{V \cdot T} \sum^{V}_{v=1} \sum^{T}_{t=1} \sum_{d} A^{v}_{d,t} \cdot L_{IOU}(\hat{B}(t,v), B(d)) 
\]

where the third sum is taken over all detections $d$ that have been associated with a ground truth annotation by Hungarian matching for that view, $B(d)$ is the bounding box of that ground truth annotation, and $L_{IOU}$ is the IOU loss. Similarly one can define $\mathcal{L}_{track\_GIOU}$ by replacing the IOU loss with the generalized IOU loss \cite{rezatofighi2019generalized}. 

While the bounding boxes predicted from the track embeddings are not as accurate as the ones predicted from the detection embeddings, the auxiliary track losses serve an important role in ensuring consistency of object identities.  

The final loss used to train the entire system end to end is:
\begin{align}
    \mathcal{L} = \, & \mathcal{L}_{det} + \mathcal{L}_{det\_aux} + \mathcal{L}_{across\_cams} + \mathcal{L}_{across\_frames} \nonumber \\
     & + \mathcal{L}_{track\_IOU} + \mathcal{L}_{track\_GIOU} \nonumber
\end{align}

\subsection{Training Protocol}
\label{sec:training_protocol}

The training proceeds based on contiguous video segments which are sampled randomly from the training data. The video segments are split into non-overlapping four frame clips, with each clip serving as a training instance. At the beginning of the video segment the track embeddings are set to an initial embedding (which is learned). For each subsequent clip, the track embeddings are initialized with the final track embeddings from the previous clip.

\begin{figure}
    \centering
        \includegraphics[width=0.8\linewidth]{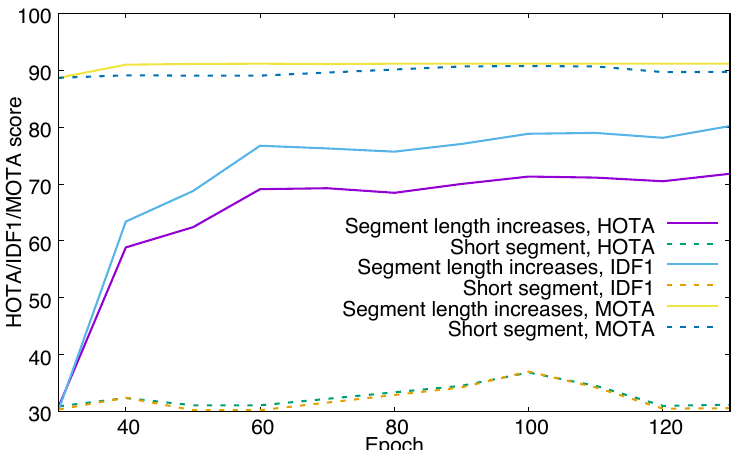}
        \caption{Effect of training protocol.  }
        \label{fig:effect_of_training_protocol}
\end{figure}

Multi-camera tracking applications usually require systems to run continuously, and track targets over time spans of minutes or tens of minutes. This creates challenges for training an end-to-end system: if training is done only over short video segments, the mismatch between training and deployment conditions introduce a domain shift that may lower the performance. On the other hand, if training is performed over long video segments, the model would see very correlated data that lacks diversity and would tend to overfit to it. To address this problem we propose the following training protocol. 

For the first thirty epochs the video segments are short, four frame clips. This stage of training is used to ensure that the model sees diverse data, which is especially important for the detector models. After this initial stage, the parameters of the detector model are frozen and the training of the tracking and association modules continues on increasingly longer video segments. The length of the video segments is randomly chosen from a geometric distribution with the expected value increasing linearly as training progresses. 

Figure~\ref{fig:effect_of_training_protocol} shows the benefit of this training protocol. As the length of the training video segments increases, performance on the test data improves, especially for the IDF1 and HOTA metrics which emphasize long term tracking accuracy. 
For MOTA score (which emphasizes detection accuracy) the performance remains relatively constant.
For comparison, if the the training video segments are kept short, the IDF1 and HOTA metrics do not improve significantly, showing that the increase in performance is due to the devised training protocol, not simply due to extra training.

\subsection{Inference}
\label{sec:prediction}

Inference on test data are made frame by frame in an online manner. In each view $v$, detections that have a predicted confidence above a threshold (0.9 for experiments in this paper) are kept and the rest are discarded. Then the Hungarian algorithm, with predicted association matrix $A^{v}$ as the weight function, is used to find a bipartite matching between detections that have not been discarded and tracks. This aims to associate detections with tracks in a way that maximizes the total association probability. If a track has not been associated with any detection for more than four frames, its embedding is reset to the initial embedding. All other track embeddings remain the same. 

\begin{figure}
    \centering
        \includegraphics[width=0.8\linewidth]{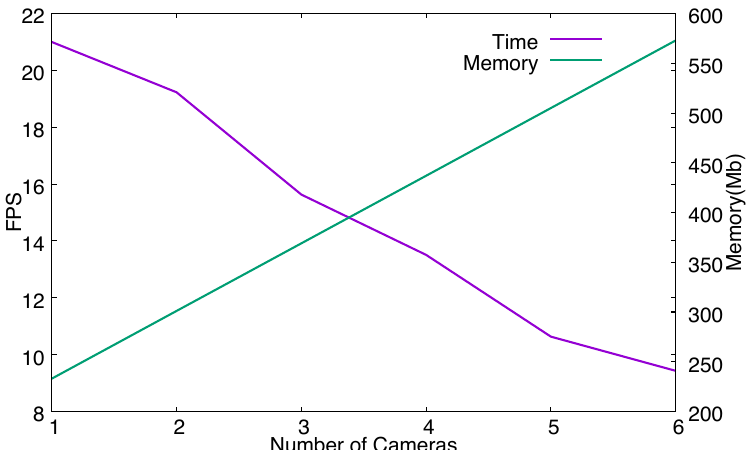}
        \caption{Inference time compute resources.}
        \label{fig:compute_resources}
\end{figure}

Figure~\ref{fig:compute_resources} shows the memory and time MCTR uses for inference as a function of the number of cameras. Both inference time and memory scale linearly with the number of cameras. With 1 camera, the model has about 44M parameters and takes 233Mb of memory and can run at 21 FPS on a single GeForce RTX 2080 Ti GPU. Each additional camera adds 1M parameters, 68Mb of memory and reduces the FPS by about 2. Thus MCTR would be well suited for real-time multi-camera tracking applications.

\section{Experimental Results}
\label{sec:results}

We contrast MCTR with the single-camera end-to-end tracker MOTR \cite{zeng2022motr}, when the later is applied to each camera view independently. \footnote{While the performance of MOTR may be improved by using pre-trained object detectors to generate detection queries, as in MOTR v2 \cite{zhang2023motrv2}, the same techniques are equally applicable to MCTR. In fact MCTR can be used directly with detectors such as YOLO foregoing DETR altogether. However, this would  distract from the goal of understanding the behaviour of end-to-end approaches.}  We also compare with ReST~\cite{cheng2023rest}, a recent multi-stage multi-camera tracking approach. When using a pre-trained YOLO detector ReST is unable to maintain track identities over longer time frames leading to very low performance on our evaluation datasets.  This highlights the shortcomings of multi-stage methods that require significant effort to ensure that all of the heuristic components are well tuned for the dataset. In contrast, end-to-end methods require much less manual tuning, learning to adapt to the characteristics of each dataset.  To gauge tracking performance of ReST, with an ideal detector, we also report results for ReST when using the ground truth bounding boxes as detections (ReST-GT).

\begin{figure*}
    \centering
    \includegraphics[width=.3\linewidth]{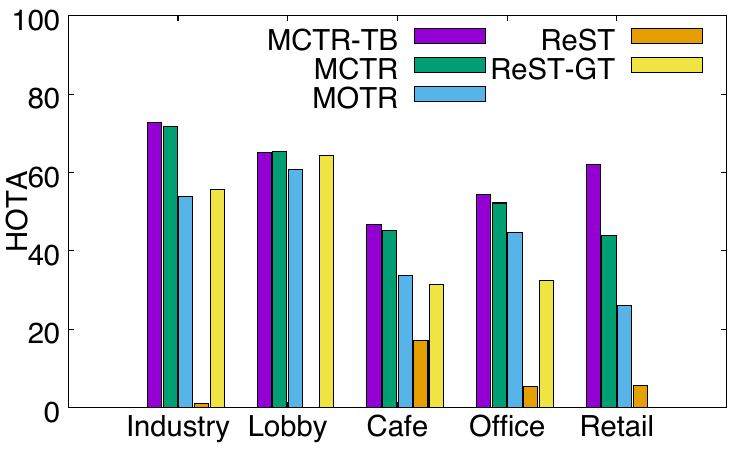}
    \includegraphics[width=.3\linewidth]{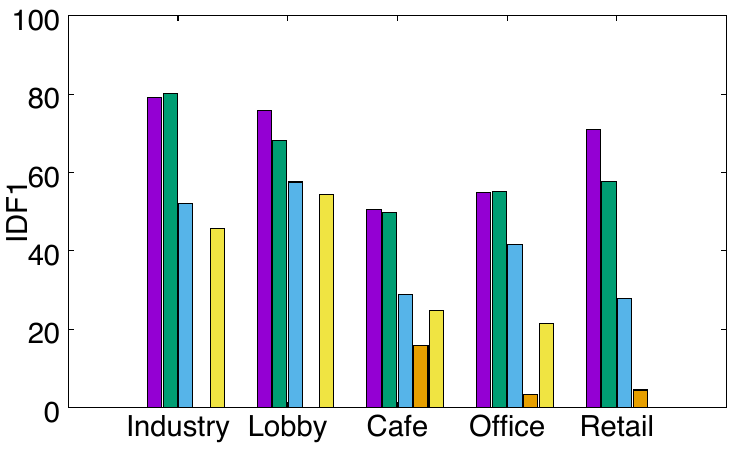}
    \includegraphics[width=.3\linewidth]{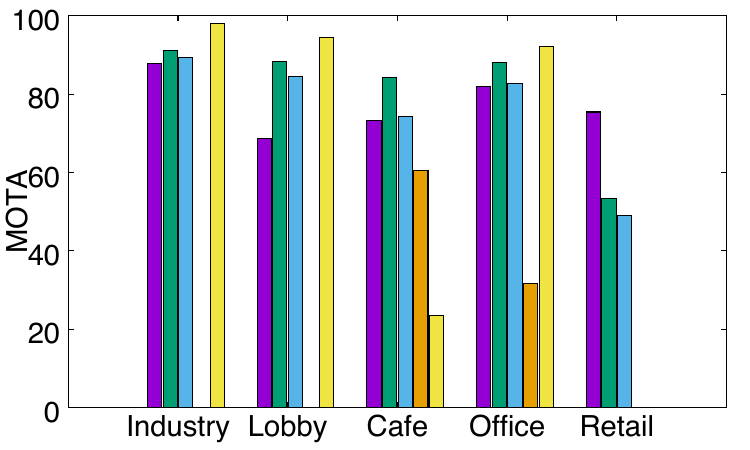} \\
    \vspace{5pt}
    \includegraphics[width=.3\linewidth]{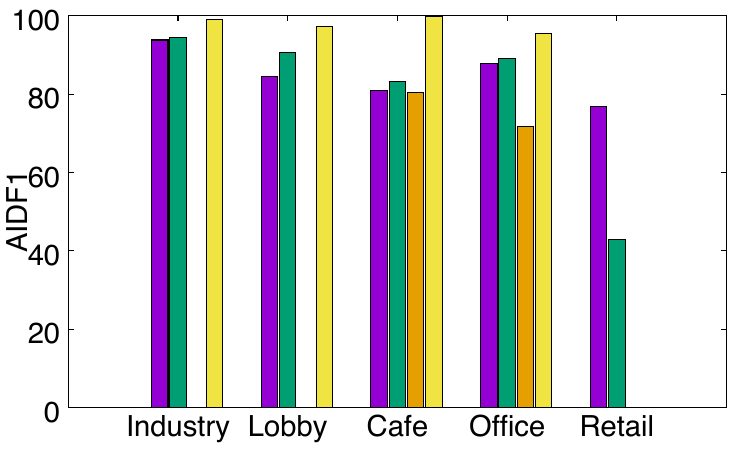}
    \includegraphics[width=.3\linewidth]{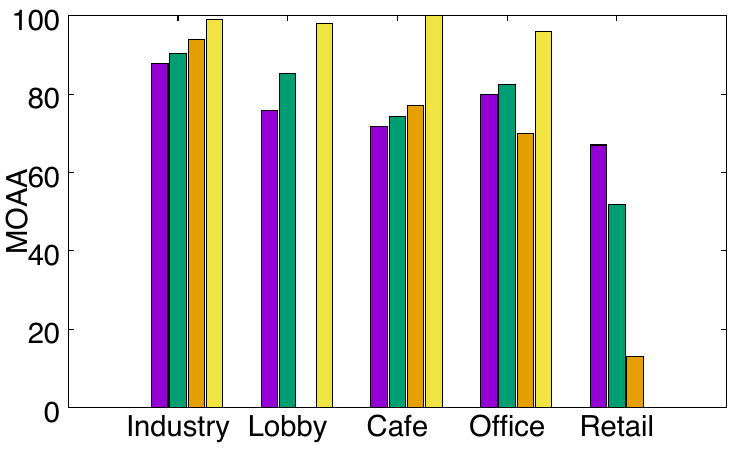}
    \caption{Results on MMPTrack dataset.  }
    \label{fig:MMPTrack}
\end{figure*}
  
MOTR is trained on each camera view and scene using the default parameters suggested by the authors. During testing we discovered an interesting failure mode for MOTR. It tends to produce large number of highly overlapping duplicate detections and consequently tracks. This adversely affects the detection precision resulting in lower scores across all metrics especially MOTA \cite{mota}. To fix the issue, we apply non-maximum suppression (NMS) and filter boxes with high overlap using an IoU threshold of 0.95. 



The models are evaluated using popular metrics in single-camera multi-object tracking literature: HOTA\cite{hota}, IDF1\cite{idf1} and MOTA\cite{mota}. The metrics are computed independently for each camera view and test clip, and the results are averaged to get the final performance for each environment. This is similar to the protocol used in the MMP-Tracking challenge \cite{mmptrackchallenge}. For MCTR and ReST we evaluate the cross-view association performance using the AIDF1 and MOAA metrics as proposed in \cite{AIDF1_2020, AIDF1_2021, CVAA}. AIDF1 is the association $F_1$ score calculated as the geometric mean of precision and recall of predicted pairwise object associations averaged across all frames and all pairs of camera views. MOAA (Multi Object Association Accuracy) follows the calculation of MOTA, but ID switches are measured across all pairs of views rather than across frames.


\subsection{MMPTrack dataset}

MMPTrack~\cite{han2023mmptrack} is a large scale multi-camera multi-person tracking dataset comprised of five environments: industry, retail, cafe, lobby and office. The data is densely labeled with the help of RGBD cameras and verified and corrected by human labelers. 

One artifact of the way the MMPTrack annotations have been collected is that ground-truth bounding boxes are given for all views, even if the person might be fully occluded in a particular view. In most environments occlusions are rare, making it less problematic. In the retail setting, however, a large fraction of people are occluded by shelves. Training a detector with these annotations would be ambiguous and lead to lower performance. To mitigate this issue, we filter out annotations of occluded persons: we employ a pre-trained YOLOX model \cite{ge2021yolox} to detect people in each frame, and ground truth annotations with a lower IOU than 0.01 with a YOLOX detection are filtered out. While this approach may eliminate annotations for visible people that are undetected by YOLOX, it serves as a reasonable approximation. The models are evaluated on the original labels, so low recall is expected in this environment.

Figure~\ref{fig:MMPTrack} shows the performance of MCTR, MOTR, ReST and ReST-GT ( ReST with ground truth bounding boxes) on the validation clips from the five MMPTrack environments \footnote{ReST-GT inference did not finish on the Retail environment, and ReST did not finish on Lobby environment}. These results represent averages across various cameras and clips. MCTR outperforms ReST and MOTR across all environments and metrics, showcasing the effectiveness of end-to-end methods for multi-camera tracking. Its higher IDF1 and HOTA scores highlight MCTR's ability to use multi-view cues to handle occlusions and maintain consistent long-term tracks.
 Compared to ReST-GT, MCTR performs better on the HOTA and IDF1 metrics, showing that even with a perfect detector, ReST temporal tracking performance is lower than MOTR. On MOTA, MOAA and AIDF1, which weight detection accuracy more heavily, ReST-GT performs better than MCTR due to its perfect detector.   



\begin{figure*}
    \centering
    \includegraphics[width=.3\linewidth]{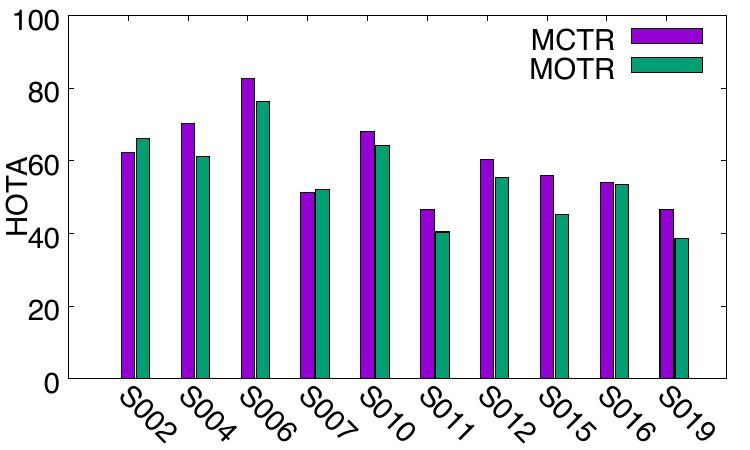}
    \includegraphics[width=.3\linewidth]{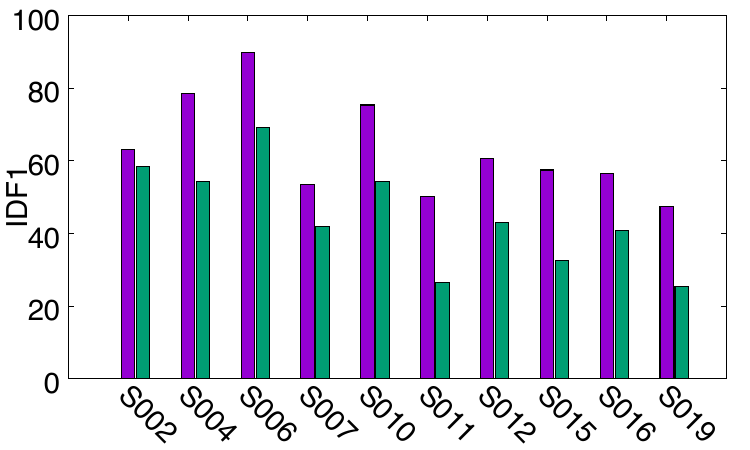}
    \includegraphics[width=.3\linewidth]{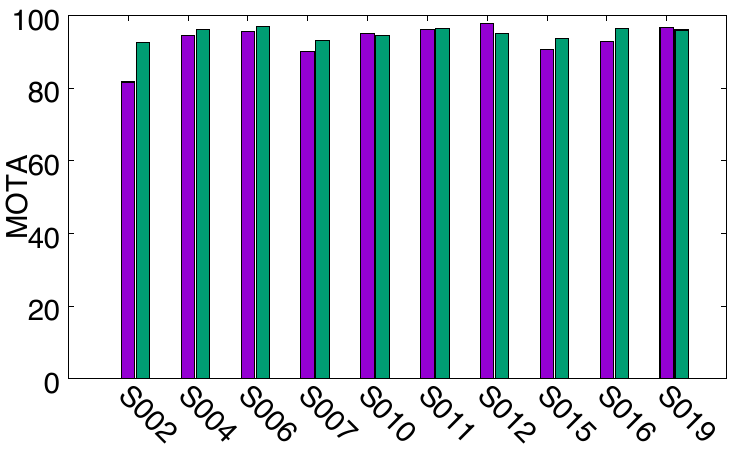} \\
    \caption{Results on AI City Challenge dataset.  }
    \label{fig:nvidia}
\end{figure*}

As mentioned above, in the retail environment there are a large number of ground-truth detections that are occluded in any particular camera view resulting in a significant number of missed detections.  While a person might be occluded in one or more camera views, it is usually visible in at least one camera allowing MCTR the opportunity to detect that person. As discussed in section~\ref{sec:loss} MCTR is trained to predict bounding boxes for all views based on the global track embeddings. While these bounding boxes are primarily utilized for computing an auxiliary loss to enhance model training, they can be also used to predict bounding boxes for occluded people that can not be detected by the view-specific detector.  When using the track-level bounding boxes for prediction (denoted as MCTR-TB in figure~\ref{fig:MMPTrack}), the performance improves significantly on the retail environment, mainly driven by an increase in recall from 54.28 to 88.73. This is remarkable, because the model has not seen any ground truth annotations in the occluded regions during training. This suggests that the model has learned an approximate geometry and is able to extrapolate it to regions where it has not seen any training data. On the other environments, the HOTA and IDF1 performance remains similar, but there is a decrease in the MOTA score due to a somewhat worse detection bounding box accuracy.

Besides the ability to predict the location of occluded people, the auxiliary losses described in section~\ref{sec:loss} are instrumental in improving the long term tracking performance of MCTR.  Table~\ref{table:aux_loss} shows the performance, on the industry environment, of MCTR with and without using the auxiliary loss. Training the model to predict detection bounding boxes from the global track embeddings leads to significant improvements in the tracking focussed HOTA and IDF1 scores, with a slight decrease in the more detection focused metrics.

\begin{table}
\caption{Performance of MCTR with and without using auxiliary losses for training}
\centering
\setlength{\tabcolsep}{3pt}
\begin{tabular}{|l|c|c|c|c|c|c|}
\hline
Model      & \small HOTA  & \small IDF1 & \small MOTA & \small AIDF1 & \small MOAA  \\ 
\hline
MCTR       & 71.81   & 80.21   & 91.19  & 94.52 &    90.39    \\ 
MCTR-noaux  & 68.00    & 75.26  & 92.11  & 95.00 &    91.30   \\ 
\hline
\end{tabular}

\label{table:aux_loss}
\end{table}

\begin{table}
\caption{Comparison with top methods in the MMP-Tracking challenge, camera view evaluation. The ranks MCTR would have achieved for each metric are shown in parentheses.}
\centering
\setlength{\tabcolsep}{3pt} 
\resizebox{\linewidth}{!}{ 
\begin{tabular}{|l|c|c|c|c|c|}
\hline
Model     & \small IDF1  & \small MOTA & \small Online & \small Camera parameters & \small FPS  \\ 
\hline
Hikvision & 86.36    & 87             & No  & Yes &    -        \\ 
Alibaba   & 88.18    & 78             & No  & Yes &    $<1$     \\ 
MCTR      & 62.30 (5) & 81.01 (3)      & Yes &  No &    9        \\ 
\hline
\end{tabular}
}
\label{table:sota-comp}
\end{table}

To the best of our knowledge, state of the art performance on the MMPTrack dataset has been achieved during the MMP-Tracking challenge\cite{mmptrackchallenge}. The highest  MOTA score on the leaderboard is 87, averaged across all five environments, and the highest IDF1 score is 88.18 \cite{mmptrackchallengeresults}. Besides using a plethora of engineered heuristics,  both these methods are offline (i.e. they can use information from the future), and make heavy use of camera calibration parameters. MCTR is online, and does not use camera parameters. With a combined MOTA of 81.01 and a combined IDF1 of 62.30, MCTR would rank thrid in MOTA and fifth in IDF1.\footnote{Our performance is measured on the validation set as we do not have access to the test set. We assume that validation and test performances are comparable.}  While the MOTA performance is respectable, there is a large gap in the IDF1 performance. This indicates that maintaining consistent tracking over a long period of time is still challenging for an end-to-end method.  

\subsection{AI City Challenge dataset}

We also evaluate on data from 2023 AI City Challenge \cite{Naphade23AIC23} Track 1, a synthetic dataset containing several indoor scenes with multiple overlapping cameras. Because our approach requires the same camera setup for training and test we deviate from the challenge setup, and instead use the first 70\% of the clip for each scene for training and the rest of 30\% for testing. We down-sample the resolution to 640x360 and use a 15 FPS frame rate.

Figure~\ref{fig:nvidia} shows the results for MCTR and MOTR on the AI City Challenge dataset. \footnote{We do not run ReST on this dataset because of its poor performance on MMPTrack, and the lack of ground truth camera calibration parameters which are required by ReST.}   MCTR again outperforms MOTR on IDF1 and HOTA metrics indicating improved long-term tracking and robustness to occlusions.  MOTR on the other hand tends to have better MOTA score, mainly due to a higher detection precision and recall. The improved detection performance of MOTR is likely due to their use of deformable DETR \cite{zhu2021deformable} which is known to perform better than standard DETR at detecting small objects, and to a prevalence of highly occluded persons (e.g. only head is visible behind shelves) in the dataset leading to a significant number of small objects.

\section{Conclusion \& Limitations}

In this paper we presented the Multi-Camera Tracking tRansformer (MCTR) a novel architecture that integrates detection and tracking across multiple cameras with overlapping fields of view in one coherent, end-to-end trainable system. MCTR is build on three key ideas: using separate track and detection embeddings, learning a probabilistic association between tracks and detections, and using a loss function based on the model's prediction of pairs of detections belonging to the same track. MCTR outperforms end-to-end single camera approaches on the MMPTrack and AI City Challenge datasets,  generating robust tracks that are consistent across time as well as camera views.

Compared to highly optimized heuristic systems that led the MMP-Tracking competition, MCTR does lag behind, particularly in IDF1 score. The lower long-term performance may stem from MCTR's Markovian approach, where track embeddings are updated only with current and previous frame information. Early attempts to use longer temporal history did not improve results. Finding better ways to integrate more temporal and motion information could enhance long-term tracking performance.

Another limitation of MCTR, and of end-to-end systems in general, is the need for extensive labeled data for training. Moreover, MCTR's reliance on camera-view specific cross-attention modules means it is closely tied to the particular camera setup used during training. An interesting question is whether a more general, camera setup independent, end-to-end model can be devised. This could involve using camera calibration parameters to adapt a single cross-attention module for all views, rather than separate modules. We leave this direction for future work.

{\small
\bibliographystyle{ieee_fullname}
\bibliography{main}
}



\end{document}